\documentclass[conference]{IEEEtran}
\IEEEoverridecommandlockouts

\usepackage{cite}
\usepackage{amsmath,amssymb,amsfonts}
\usepackage{algorithmic}
\usepackage{graphicx}
\usepackage{textcomp}
\usepackage{xcolor}

\makeatletter
\newcommand{\linebreakand}{%
  \end{@IEEEauthorhalign}
  \hfill\mbox{}\par
  \mbox{}\hfill\begin{@IEEEauthorhalign}
}
\makeatother

\usepackage{url}
\usepackage{booktabs}
\usepackage{multirow}

\def\BibTeX{{\rm B\kern-.05em{\sc i\kern-.025em b}\kern-.08em
    T\kern-.1667em\lower.7ex\hbox{E}\kern-.125emX}}
\begin{document}

\title{Generate E-commerce Product Background \\ by Integrating Category Commonality \\ and Personalized Style
}

\author{\IEEEauthorblockN{Haohan Wang}
\IEEEauthorblockA{\textit{JD.COM}\\
Beijing, China \\
wanghaohan1@jd.com}
\and
\IEEEauthorblockN{Wei Feng}
\IEEEauthorblockA{\textit{JD.COM}\\
Beijing, China \\
fengwei25@jd.com}
\and
\IEEEauthorblockN{Yaoyu Li}
\IEEEauthorblockA{\textit{JD.COM}\\
Beijing, China \\
liyaoyu1@jd.com}
\and
\IEEEauthorblockN{Zheng Zhang}
\IEEEauthorblockA{\textit{JD.COM}\\
Beijing, China \\
zhangzheng11@jd.com}
\linebreakand
\IEEEauthorblockN{Jingjing Lv}
\IEEEauthorblockA{\textit{JD.COM}\\
Beijing, China \\
lvjingjing1@jd.com}
\and
\IEEEauthorblockN{Junjie Shen}
\IEEEauthorblockA{\textit{JD.COM}\\
Beijing, China \\
shenjunjie@jd.com}
\and
\IEEEauthorblockN{Zhangang Lin}
\IEEEauthorblockA{\textit{JD.COM}\\
Beijing, China \\
linzhangang@jd.com}
\and
\IEEEauthorblockN{Jingping Shao}
\IEEEauthorblockA{\textit{JD.COM}\\
Beijing, China \\
shaojingping@jd.com}
}

\maketitle

\begin{abstract}
The state-of-the-art methods for e-commerce product background generation suffer from the inefficiency of designing product-wise prompts when scaling up the production, as well as the ineffectiveness of describing fine-grained styles when customizing personalized backgrounds for some specific brands. To address these obstacles, we integrate the category commonality and personalized style into diffusion models. Concretely, we propose a Category-Wise Generator to enable large-scale background generation with only one model for the first time. A unique identifier in the prompt is assigned to each category, whose attention is located on the background by a mask-guided cross attention layer to learn the category-wise style. Furthermore, for products with specific and fine-grained requirements in layout, elements, etc, a Personality-Wise Generator is devised to learn such personalized style directly from a reference image to resolve textual ambiguities, and is trained in a self-supervised manner for more efficient training data usage. To advance research in this field, the first large-scale e-commerce product background generation dataset BG60k is constructed, which covers more than 60k product images from over 2k categories. Experiments demonstrate that our method could generate high-quality backgrounds for different categories, and maintain the personalized background style of reference images. BG60k will be available at \url{https://github.com/Whileherham/BG60k}.
\end{abstract}


\section{Introduction}
E-commerce product background generation is practically valuable in sparking the purchase desires. The natural and realistic backgrounds generated turn out to exhibit better online performance than the simple ones \cite{bhamidipati2017large,zhou2019understanding,mishra2020learning,8622259,fan2022automatic,belem2019image,natadirja2023commerce,Jiang_2022_CVPR,zhendong2022make,li2014impact,pereira2022influence,nitse2004impact}, \textit{e.g.}, around $20\%$ higher click-through rate (CTR) in JD.COM. Due to the significant differences in the background of advertising images from different categories and brands, advertisers traditionally employ professional designers to design suitable backgrounds for their products, which is often time-consuming and financially burdensome.

\begin{figure}[t]
	\centering
	\includegraphics[width=0.7\columnwidth]{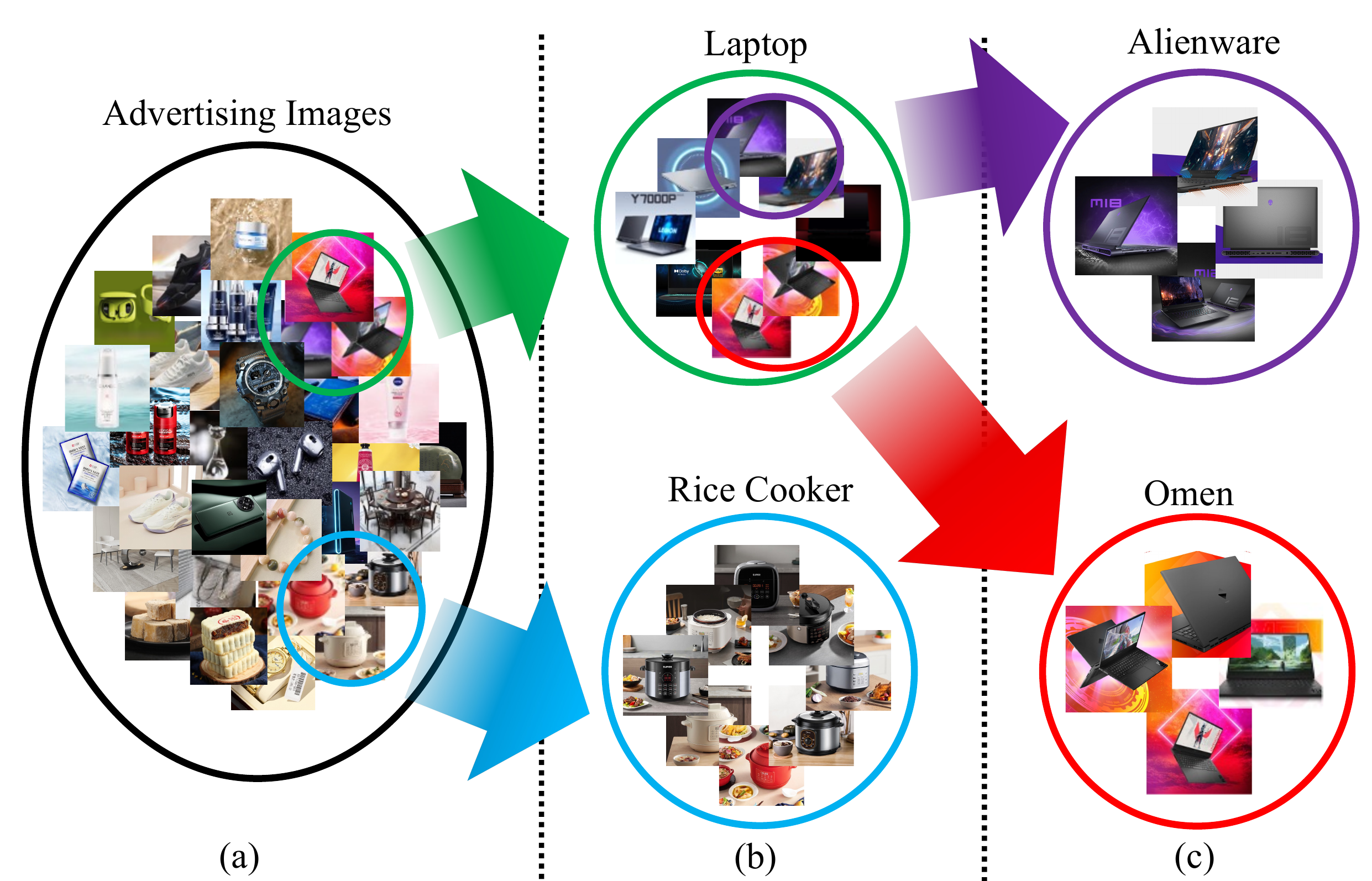}
	\caption{(a) Large-scale advertising images from the e-commerce platform. (b) Advertising images clustered by the category. (c) Advertising images clustered by the brand. Images of some brands have their signature styles.}
	\label{fig:department-personalized}
 \vspace{-2.0em}
\end{figure}

To reduce this cost, most previous methods \cite{mishra2020learning,chen2021efficient,chen2021automated,wei2022towards} pre-generate a template image, which the product is then pasted on. As the background is independent of the product during generation, the final image exhibits a discernibly low level of realism. To address this issue, recently text-based inpainting methods \cite{suvorov2022resolution,ldm,bd,bld,controlnet,glide,dalle2,wang2023imagen,sdxl,mutual,yu2023inpaint} are widely applied, where the background is described by several sentences (prompt) and generated with the consideration of the product. Though not designed specifically for e-commerce product, they surpass the template-based methods in aesthetics. However, it is still time-consuming to design proper prompt for each product to describe the content of the background, harming the efficiency in large-scale background generation, as shown in Fig. \ref{fig:department-personalized} (a). Besides, products of some specific brands demand backgrounds in a fine-grained and consistent personalized style, which is ineffective to describe only by text.

For achieving large-scale background generation efficiently, we integrate the category commonality into diffusion models. We observe that the intra-category variance among the background styles is significantly less than the inter-category variance. As shown in Fig. \ref{fig:department-personalized} (b), the backgrounds of laptops usually protrude a sense of modern technology, while the rice cookers are mainly on the dining tables. Hence we present a Category-Wise Generator, where we templatize the prompt by assigning an identifier to each category for its common background style. A mask-guided cross attention layer is also devised to enforce the attention of identifier only on the background. After training, the identifier could serve as the alternative to the category-wise prompt.


Nevertheless, beyond the common character in the category, the advertising images of some well-known brands exhibit their signature styles in layout, elements, color, lighting, etc. As shown in Fig. \ref{fig:department-personalized} (c), in addition to the sense of technology, the background of Alienware is mainly purple, while Omen is mainly red. Therefore, we further propose a Personality-Wise Generator in parallel with the Category-Wise Generator, to incorporate the personalized style into the category-wise general background. Due to the poor effectiveness of textual representation in describing fine-grained or abstract information, the Personality-Wise Generator generates personalized backgrounds by directly mimicking the characters of the reference advertising images. To leverage the training data more efficiently, we adopt the self-supervised training to consider an advertising image as both the reference and ground truth image. We also perturb the reference image by a series of augmentation strategies during training, which is crucial to prevent the generated background from a simple copy of the reference image during inference.

Considering there is no publicly available dataset for large-scale product background generation methods, we elaborately mine and clean a large amount of advertising images from JD.COM to fill in this gap. The finally released dataset covers over 2k categories and contains more than 60k images.

Our contributions are summarized as follows:
\begin{itemize}
\item
We present a Category-Wise Generator to enable the efficient large-scale background generation with only one model for the first time. Specifically, we assign an identifier to each category and design a mask-guided cross attention layer to locate the attention of the identifier only on the background. The identifier finally serves as the prompt for category-wise background style. 
\item
We propose a Personality-Wise Generator to maintain the personalized styles of some brands by mimicking the characters of reference images. Considering the lack of similar background pairs, we employ the self-supervised training with a background perturbation strategy.
\item
We release the first large-scale dataset to advance the research of the product background generation. Extensive experiments demonstrate that our method outperforms the state-of-the-art inpainting methods by a large margin in both background similarity and quality.
\end{itemize}

\section{Method}
\begin{figure}[t]
    \centering
    \includegraphics[width=\columnwidth]{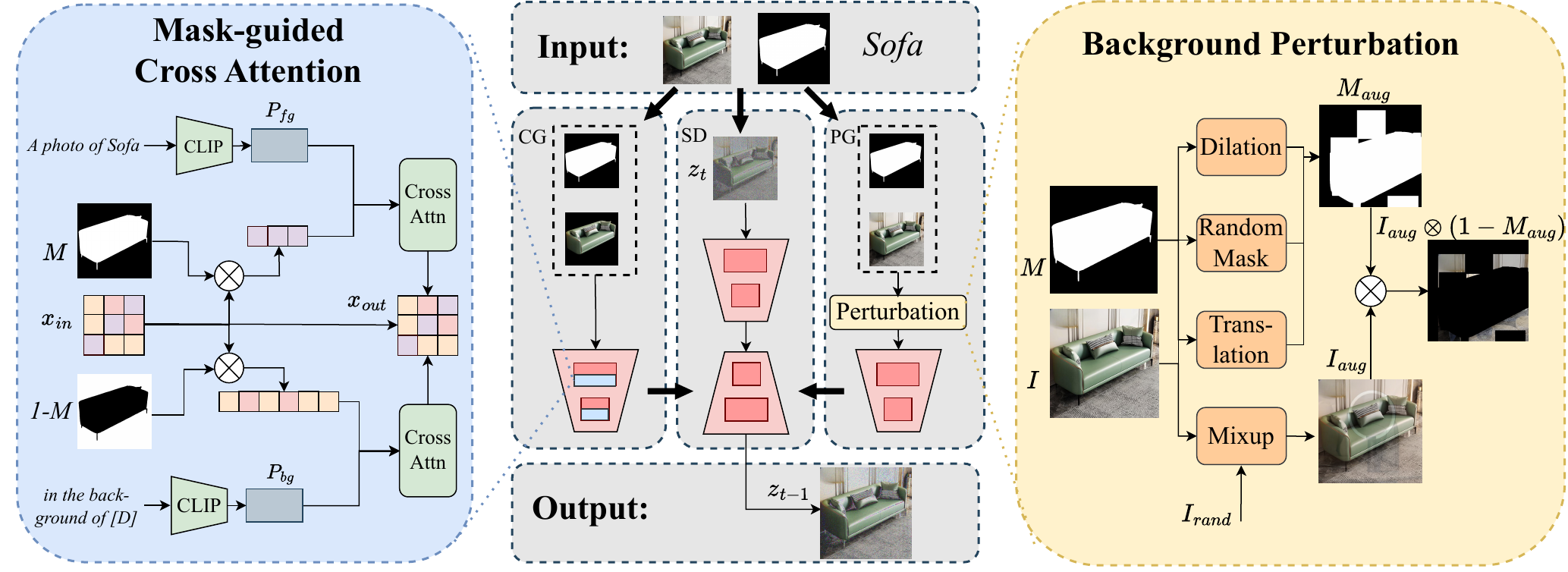}
    \caption{The overall framework of our methods during training. The noise added to the input of CG and PG is omitted for simplicity. During inference, the perturbation is removed. Notably, CG could work solely without PG.
    }
    \label{fig:pipeline}
    \vspace{-1.5em}
\end{figure}

We generate the background by inpainting to ensure the authenticity of the product and the overall harmony of the image. The overall framework is illustrated in Fig. \ref{fig:pipeline}, which consists of three components: Stable Diffusion (SD), the Category-Wise Generator (CG) and the Personality-Wise Generator (PG), where the latter two modules are built and modified based on the architectures of ControlNet \cite{controlnet}. During training, given the advertising image $I$ and the mask of the product $M$, CG takes as the input $I \otimes M$ to integrate the general category knowledge (Sec. \ref{sec:Department-Knowledge}), while PG captures the personalized style from $I \otimes (1-M)$ (Sec. \ref{sec:Personalized-Style}), where $\otimes$ is Hadamard product. Then the multi-level outputs of the two generators are all incorporated into the SD Decoder in a similar way to ControlNet to predict the latent noise.

\subsection{Generation with Category Commonality}
\label{sec:Department-Knowledge}

As the backgrounds of the products in the same category have many characters in common, it is natural to consider sharing the same prompt among the products in the same categories, \textit{e.g.}, injecting the category name into a prompt template \textit{'A photo of [category]'} and inheriting the original architecture of ControlNet for inpainting. However, such naive solution is unable to distinguish the background commonality from the foreground commonality. To resolve this problem, we divide the prompt into the product prompt (\textit{'A photo of [category]'}) and the background prompt (\textit{'in the background of [D]'}). Inspired by Dreambooth \cite{dreambooth}, we map the category-wise background commonality into an identifier [D]. After that,  we devise a mask-guided cross attention layer to ensure that the product and background generation solely rely on their corresponding prompt respectively. The mask-guided cross attention layer is formulated as:
\begin{equation}
x_{out}^i = x_{in}^i + CA(x_{in}^i, P_{fg}) \otimes M + CA(x_{in}^i, P_{bg}) \otimes (1-M),
\end{equation}
where $CA(\cdot)$, $x_{in}^i$ and $x_{out}^i$ are the cross attention layer in diffusioon models, the input and output of the $i_{th}$ cross attention layer, respectively. $M$, $P_{fg}$, $P_{bg}$ indicate the mask of the product, the encoded product prompt and the background prompt, respectively.  We replace all the attention layers in CG with the mask-guided ones, forcing the sub-network to generate the background based on the background prompt only. After training, \textit{[D]} could serve as the prompt to describe the common background style of a specific category.

\subsection{Generation with Personalized Style}
\label{sec:Personalized-Style}

The backgrounds containing category-wise commonality are appropriate for most products, but not for some well-known brands, which require their consistent signature background styles. Hence superposed on the category-wise background, we further propose PG to generate the personalized background similar to the reference image in the layout, elements, etc. We also take the similar architecture of ControlNet in PG, as the output high resolution feature maps (rather than the single token \cite{paint-by-example}) of ControlNet preserve the semantic cues with spatial information naturally. As there lack enough training pairs comprising a reference image and a similar ground-truth generated image, PG is trained in a self-supervised fashion. During training, we sample an advertising image in each iteration and obtain its product and background. The product and the background are fed to CG and PG to reconstruct the image itself. In this way, the ground truth of the generated background is the original background itself. To prevent PG from learning the copy-and-paste as a shortcut, we further adopt a background perturbation strategy before the input of PG. Concretely, we firstly dilate $M$ to $M_{aug}$ to tackle the difference between the original product and the product in the reference image by inpainting the region $I \otimes (M_{aug}-M)$. Then we leverage the mixup strategy to overlay a random advertising image $I_{rand}$ onto the reference image $I$, to penalize the PG for simply replicating the elements from the reference image. Specifically, $I$ is perturbed by:
\begin{equation}
I_{aug} = \sigma \otimes I + (1-\sigma) \otimes I_{rand}, 
\end{equation}
where $\sigma$ is the hyper-parameter. We feed $I_{aug} \otimes (1-M_{aug})$ to PG instead. It turns out that PG learns to understand more about the elements in the reference image, filtering out the inappropriate portion and generating semantically similar elements for the other portion.

Although the background perturbation strategy prevents PG from the copy-and-paste, we experimentally find that it has a negative impact on the quality of the generated images. To resolve this obstacle, we utilize the original high-quality reference image as the initial latent image during inference, \textit{i.e.}, replacing $z_T$ with the latent reference image with noise $\epsilon\sim\mathcal{N}(0,1)$ added for $T$ times. This reduces the difficulty of generating images similar to reference images, and is thereby beneficial for improving the image quality.

\section{BG60k}
\label{sec:BG60K}

\begin{figure}[t]
    \centering
    \includegraphics[width=0.8\columnwidth]{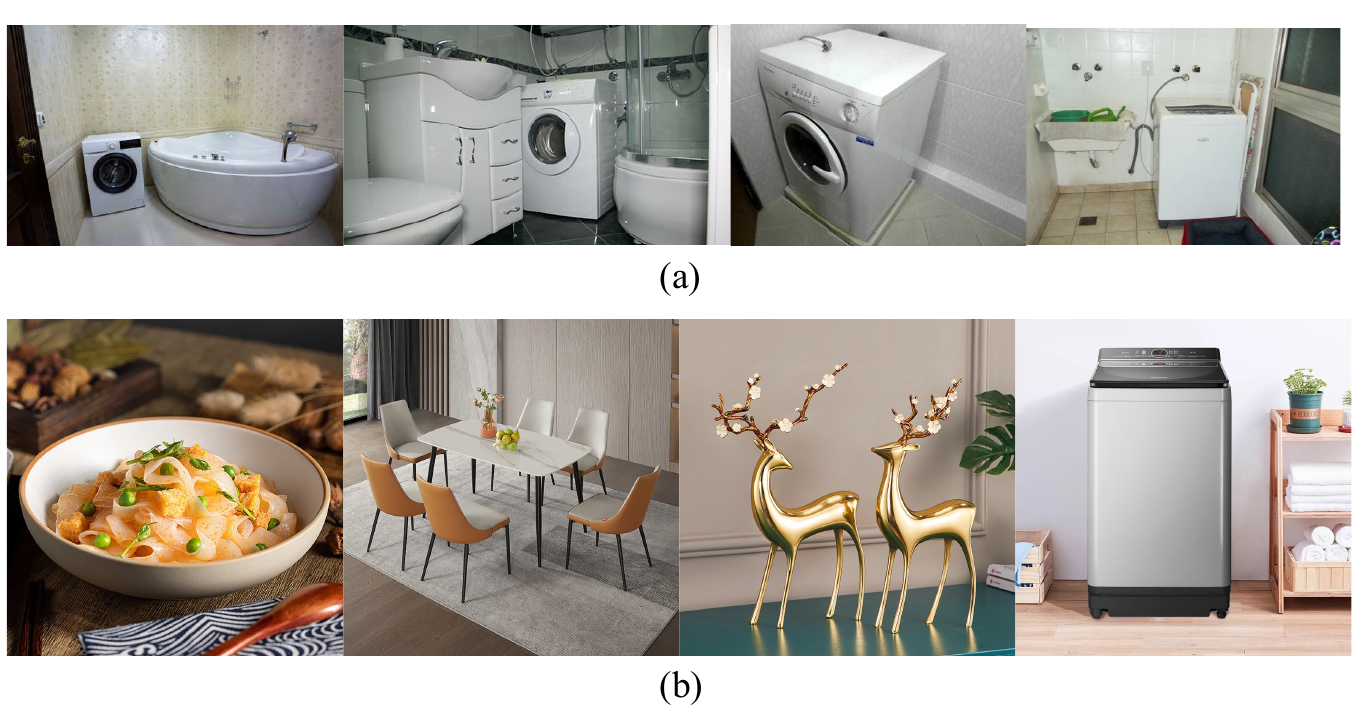}
    \vspace{-0.5em}
    \caption{(a) Some unsatisfying examples in LAION-5B \cite{laion5b}. (b) Examples in BG60k, where the training images are more appealing and exquisite.
    }
    \label{fig:laion}
    \vspace{-1.0em}
\end{figure}

It is commonly known that the performance of the generation models relies heavily on the quality of the training data. The most widely-used publicly available dataset, like LAION-5B \cite{laion5b} and its subsets, are not designed specifically for e-commerce scenarios. Hence several issues arise in the training images, such as non-salient foreground, the improper perspective, or the dirty and dusky background, as shown in Fig. \ref{fig:laion} (a). Models trained on such dirty data will fail to generate satisfying advertising images, as shown in Fig. \ref{fig:tongshi} (b). 

To reduce this gap, we release the first large-scale e-commerce background generation dataset BG60k. We first mine millions of advertising images from the famous e-commerce platform JD.COM, then clean them by removing images containing texts or with low CTR. This ends up with 63293 advertising images from 2032 categories. Each item in the dataset is composed of an original advertising image $I$, a mask of the product $M$ and the category of the product. Examples of BG60K are illustrated in Fig. \ref{fig:laion} (b). Moreover, we choose 1k products and their original backgrounds, namely BG1k, to compare the difference between the generated and original background statistically. We also select 1.6k pairs of product and reference image to evaluate the ability of generating with personalized style, namely BG-pair. 

\section{Experiments}

\subsection{Implementation Details and Evaluation Metrics}
Our model is trained on BG60k for 300 epochs. The batch size is 128 and each image is resized to $512\times512$,. The $\sigma$ in mixup is sampled from a uniform distribution $\mathcal{U}(0.75, 0.95)$. 

Similar to prior works, we adopt the FID score \cite{fid} and the CLIP similarity between generated and original advertising images to measure the quality of the generated images. Besides, a user study is conducted for a more comprehensive comparison, where fifty experienced advertising practitioners are invited to rank the overall aesthetics of generated images and the background consistency.

\subsection{Comparison with State-of-the-Art}
\noindent \textbf{Large-Scale Background Generation}
We remove PG and only evaluate CG as no reference images included. As summarized in Table \ref{tab:hanyetongshi}, CG has the superiority of 1.25 CLIP similarity and 1.85 FID. Thus the backgrounds generated by CG are more like the original high-quality backgrounds and thus better meet the category-wise requirements of the advertisements.

\begin{figure}[t]
    \centering
    \includegraphics[width=0.9\columnwidth]{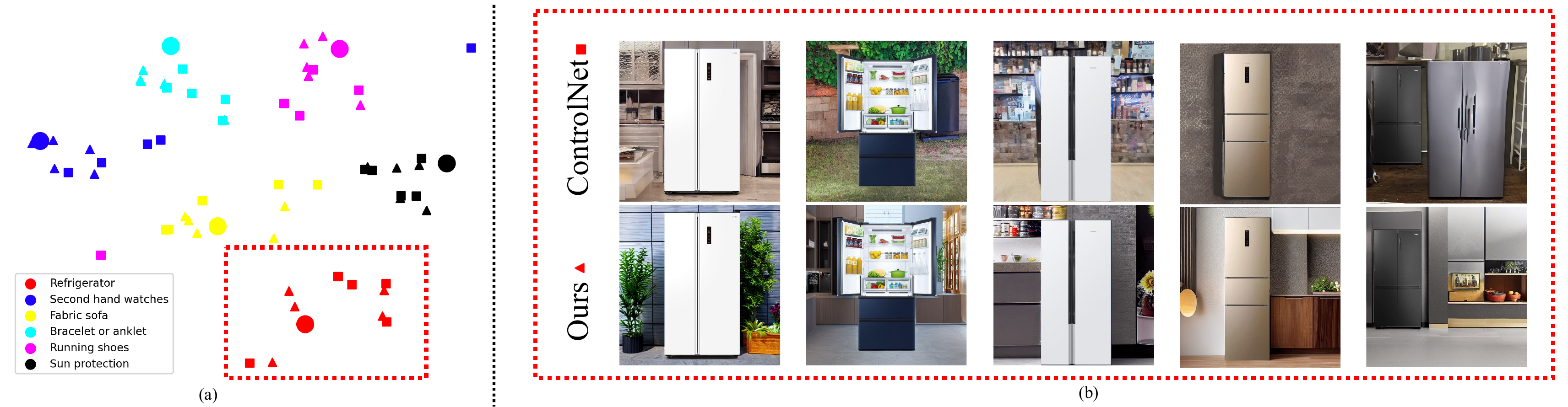}
    \caption{
    (a) tSNE visualization, where the circle represents the cluster center. Each triangle/square indicates the embedding of one image generated by CG/ControlNet. (b) Visualization of comparison on the category \textit{refrigerator}.}
    \vspace{-1.0em}
    \label{fig:tongshi}
\end{figure}

Furthermore, we visualize the features of backgrounds from CG and ControlNet by tSNE in Fig. \ref{fig:tongshi} (a). It could be seen that the features from CG are clustered around the corresponding centers more compactly. Then taking 'refrigerator' as an example, we compare the generated images in Fig. \ref{fig:tongshi} (b). It turns out that ControlNet might generate backgrounds lacking photorealism or containing other product of the same category, while CG is able to generate indoor backgrounds consistently.

\begin{table}[h]
\centering
\small
\caption{Comparison conditioned on the prompt template}
\begin{tabular}{@{}lcc@{}}
\toprule
Method & CLIP Similarity ($\uparrow$) & FID ($\downarrow$) \\ \midrule
Template-based & 86.93 & 7.93 \\
LaMa \cite{lama} & 88.33 & 8.24 \\
Stable Diffusion \cite{ldm} & 76.47 & 7.54 \\
ControlNet \cite{controlnet}& 84.80 & 5.79 \\
Ours & \textbf{89.58} & \textbf{3.94} \\ \bottomrule
\end{tabular}
\vspace{-1.0em}
\label{tab:hanyetongshi}
\end{table}

\noindent \textbf{Personalized Background Generation}
For each pair of product and reference advertising image from BG-pair, we compare our generated backgrounds conditioned on the reference image with those from the state-of-the-art image-based inpainters \cite{bd, paint-by-example,ldm,controlnet} in Table \ref{tab:personal}\footnote{$^\dagger$ and $^\ddagger$ mean initializing $z_T$ with the noisy latent reference image}. The significant superiority of CLIP similarity indicates that our method could better mimic the character of the reference image, while the extremely low FID demonstrates that our method generates backgrounds consistently following the distribution of the advertising images.

We further conduct a user study for a more focused and comprehensive comparison. We average the rank scores (five stands for the best and one for worst) and show the results in Table \ref{tab:usr}. From the participant’s subjective perspective, almost every time the background generated by our method is picked as the most similar one. Moreover, our method also ranks first in quality. This is more than challenging as pursuing the similarity in the layout and elements might hurt the harmony.

\begin{table}[h]
\centering
\caption{Background generation conditioned on the reference image}
\small
\begin{tabular}{@{}lcc@{}}
\toprule
\multirow{2}{*}{Method} & \multicolumn{2}{c}{Ref $\rightarrow$ Self} \\ \cmidrule(l){2-3}
 & CLIP Similarity ($\uparrow$)& FID ($\downarrow$) \\ \midrule
Blended Diffusion \cite{bd} & 63.51 & 3.62 \\
Paint-by-Example \cite{paint-by-example} & 62.82 & 5.96\\
Stable Diffusion$^\dagger$  \cite{bld} & 57.10 & 7.92 \\
ControlNet$^\ddagger$ \cite{controlnet}& 62.78 & 3.92 \\
Ours & \textbf{68.26} & \textbf{2.39} \\ \bottomrule
\end{tabular}
\vspace{-1em}
\label{tab:personal}
\end{table}

\begin{table}[h]
\centering
\small
\caption{User study on background similarity and image quality. }
\begin{tabular}{@{}lcc@{}}
\toprule
Method & Similarity ($\uparrow$) & Quality ($\uparrow$) \\ \midrule
ControlNet$^\ddagger$ \cite{controlnet}& 2.39 & 2.65 \\
Blended Diffusion \cite{bd} & 2.49 & 2.71 \\
Paint-by-Example \cite{paint-by-example}& 2.05 & 2.23 \\
Ours w/o PG  & 3.16 & 3.49 \\
Ours & \textbf{4.88} & \textbf{3.90} \\ \bottomrule
\end{tabular}
\vspace{-0.5em}
\label{tab:usr}
\end{table}

\begin{table}[h]
\caption{Ablation study on different modules of our method.}
\resizebox{\columnwidth}{!}{
\centering
\begin{tabular}{@{}lccccc@{}}
\toprule
\multirow{2}{*}{} & \multirow{2}{*}{\begin{tabular}[c]{@{}c@{}}Category-Wise \\ Generator\end{tabular}} & \multirow{2}{*}{\begin{tabular}[c]{@{}c@{}}Personality-Wise\\ Generator\end{tabular}} & \multirow{2}{*}{\begin{tabular}[c]{@{}c@{}}Initial\\ Image\end{tabular}} & \multicolumn{2}{c}{Ref $\rightarrow$ Self} \\ \cmidrule(l){5-6} 
 &  &  &  & Sim ($\uparrow$) & FID ($\downarrow$)\\ \midrule
(a) &  & $\checkmark$ & $\checkmark$ & 66.17 & 4.01 \\
(b) & $\checkmark$ &  & $\checkmark$ & 67.50 & 2.76 \\
(c) & $\checkmark$ & $\checkmark$ &  & 66.51 & 2.73 \\
(d) & $\checkmark$ & $\checkmark$ & $\checkmark$ & \textbf{68.26} & \textbf{2.39} \\ \bottomrule
\end{tabular}
}
\vspace{-1em}
\label{tab:aba1}
\end{table}

\subsection{Ablation Study}
\noindent \textbf{Effectiveness of CG:} We replace CG with the original ControlNet inpainting model and then a significant performance deterioration is observed from line (a) and (d) in Table \ref{tab:aba1}. It shows the necessity of the category commonality even with a reference image given, as it forms the basis of the background.

\noindent \textbf{Effectiveness of PG:} We remove PG and initialize $z_T$ with the latent reference image with noise $\epsilon\sim\mathcal{N}(0,1)$ added for $T$ times. From line (b) and (d) in Table \ref{tab:aba1}, this results in the drop of CLIP similarity by $0.76\%$, and the increase of FID by 0.37. It is because the model fails to recognize and leverage the background information explicitly from the reference image.

\noindent \textbf{Impact of Initial Latent Images:} We try not initializing $z_T$ with the noisy latent image, then as shown in line (c) and (d) in Table \ref{tab:aba1}, it has a negative influence on both metrics, proving the importance of the proper initialization o image quality.

\begin{figure}[h]
    \centering
    \includegraphics[width=0.8\columnwidth]{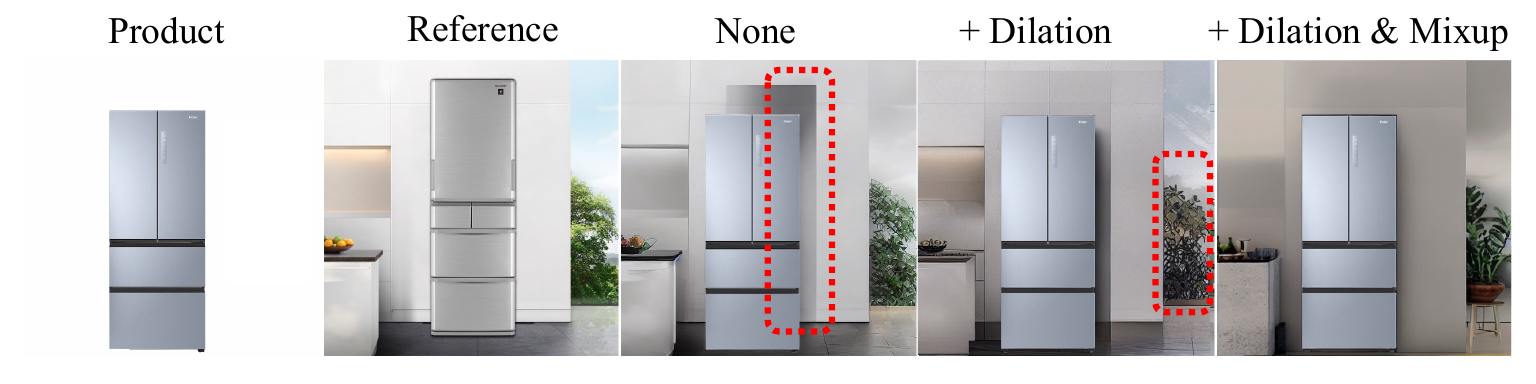}
    \vspace{-.5em}
    \caption{Comparison on different augmentation strategies.
    }
    \vspace{-.5em}
    \label{fig:ablation-aug}
\end{figure}

\noindent \textbf{Impact of Augmentation Strategies:} As illustrated in the 3rd column in Fig. \ref{fig:ablation-aug}, when no perturbation adopted, an anomaly could be observed in where the product is originally placed. With the dilation added, some elements still look like blunt copies. With the mixup strategy added, the generated background is similar but not identical to the reference background.

\begin{figure}[h]
    \centering
    \vspace{-.5em}
    \includegraphics[width=0.9\columnwidth]{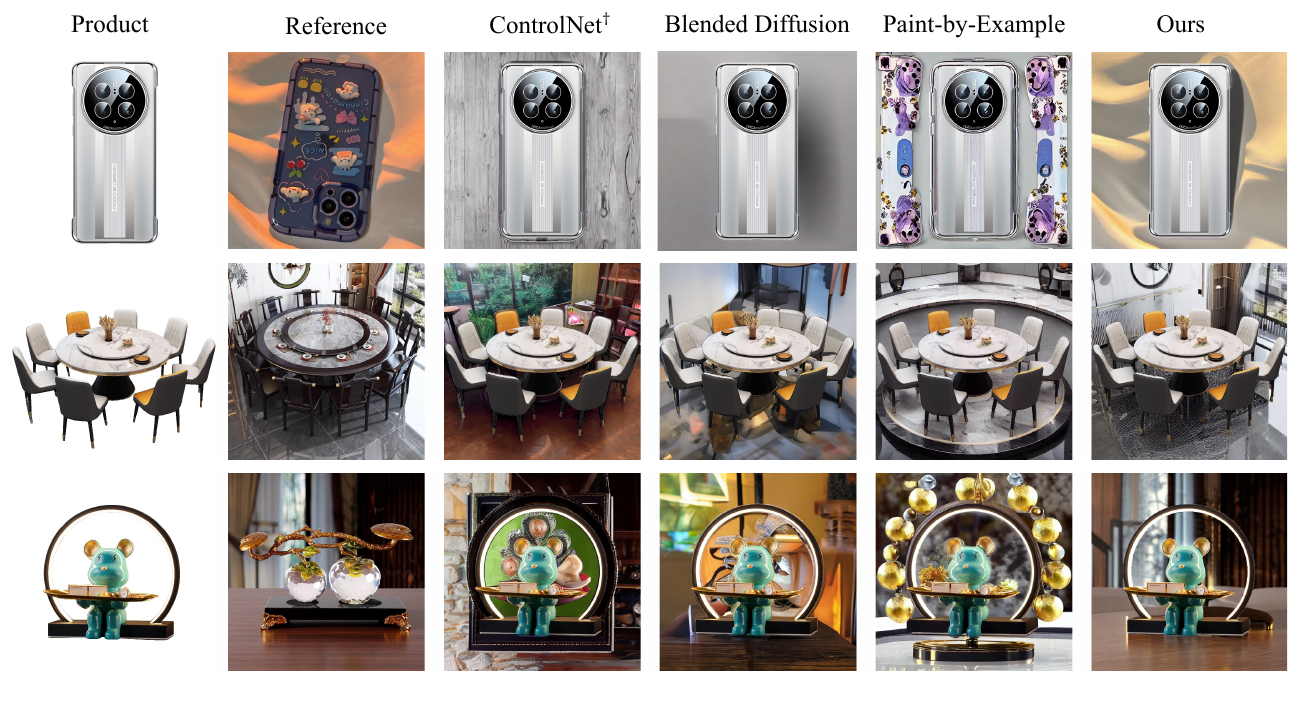}
    \vspace{-.5em}
    \caption{Qualitative Comparison. }
    
    \label{fig:qualitative-results}
    \vspace{-1.0em}
\end{figure}

\noindent \textbf{Qualitative Comparison:}
As shown in Fig. \ref{fig:qualitative-results}, our method could mimic both the abstract style (1st row) and concrete scene (the last two rows) of the reference image better. 

\section{Conclusion}
In this work, we improve the efficiency of large-scale generation by a Category-Wise Generator, and maintain the personalized styles for some specific brands from reference images effectively by a Personality-Wise Generator. We also release the first large-scale background generation dataset. Experiments show the comprehensive superiority of our methods.

{
 \bibliographystyle{IEEEtran}
 \bibliography{reference}
}

\end{document}